\documentclass{article}
\usepackage{spconf,amsmath,epsfig}
\usepackage{float}
\usepackage{multirow}
\usepackage{amssymb}
\usepackage{graphicx}

\let\OLDthebibliography\thebibliography
\renewcommand\thebibliography[1]{
  \OLDthebibliography{#1}
  \setlength{\parskip}{0pt}
  \setlength{\itemsep}{0pt plus 0.3ex}
}

\begin{document}\sloppy

\title{PYRAMID FEATURE ATTENTION NETWORK FOR MONOCULAR DEPTH PREDICTION}

\name{Yifang Xu, Chenglei Peng, Ming Li, Yang Li, and Sidan Du\thanks{Corresponding authors: Chenglei Peng and Sidan Du}}
\address{Nanjing University, Nanjing Institute of Advanced Artificial Intelligence, Nanjing, China\\  
\{mf20230128, dg20230020\}@smail.nju.edu.cn, \{pcl, yogo, coff128\}@nju.edu.cn
}

\maketitle

\begin{abstract}
Deep convolutional neural networks (DCNNs) have achieved great success in monocular depth estimation (MDE). However, few existing works take the contributions for MDE of different levels feature maps into account, leading to inaccurate spatial layout, ambiguous boundaries and discontinuous object surface in the prediction. To better tackle these problems, we propose a Pyramid Feature Attention Network (PFANet) to improve the high-level context features and low-level spatial features. In the proposed PFANet, we design a Dual-scale Channel Attention Module (DCAM) to employ channel attention in different scales, which aggregate global context and local information from the high-level feature maps. To exploit the spatial relationship of visual features, we design a Spatial Pyramid Attention Module (SPAM) which can guide the network attention to multi-scale detailed information in the low-level feature maps. Finally, we introduce scale-invariant gradient loss to increase the penalty on errors in depth-wise discontinuous regions. Experimental results show that our method outperforms state-of-the-art methods on the KITTI dataset.
\end{abstract}
\section{Introduction}
\label{sec:intro}
Monocular depth estimation (MDE) is an important task that aims to predict pixel-wise depth from a single RGB image, and has many applications in computer vision, such as 3D reconstruction, scene understanding, autonomous driving and intelligent robots \cite{DBLP:conf/nips/EigenPF14}. In the meanwhile, MDE is a technically ill-posed problem as a single image can be projected from an infinite number of different 3D scenes. To solve this inherent ambiguity, one possibility is to leverage prior auxiliary information, such as texture information, occlusion, object locations, perspective, and defocus \cite{DBLP:conf/cvpr/FuGWBT18/DORN}, but it is not easy to effectively extract useful prior information. 

More recently, some works on MDE based on encoder-decoder architecture have shown significant improvements in performance by using deep convolutional neural networks (DCNNs) \cite{DBLP:journals/corr/abs-1907-10326/BTS}. As backbone for encoder, very powerful deep networks such as ResNet \cite{DBLP:conf/cvpr/HeZRS16/ResNet}, DenseNet \cite{Huang2017DenselyCC/DenseNet} or ResNext \cite{DBLP:conf/cvpr/XieGDTH17/ResNext} are widely adopted. These networks cascade multiple convolutions and spatial pooling layers to gradually increase the receptive field and generate the high-level depth information. In decoder phase, state-of-the-art methods are based on upsampling layer with global context module \cite{DBLP:journals/corr/abs-1812-11941/DenseDepth}, skip connection, depth-to-space \cite{DBLP:journals/corr/abs-2009-00743/BANet}, multi-scale local planar guidance for upsampling operation \cite{DBLP:journals/corr/abs-1907-10326/BTS}. These methods directly fuse different scale features without considering their different contributions for MDE, which leads to ambiguous boundaries and discontinuous object surface in predicted depth (see Fig.1 (c)). To tackle these problems, logarithmic discretization for ordinal regression \cite{DBLP:conf/cvpr/FuGWBT18/DORN} and attention module with structural awareness \cite{Tian2020/kuaishou} are introduced to MDE network. However, the high-level and low-level features play different roles in MDE. The existing methods did not consider this aspect, which may affect the effective extraction of depth information.
\begin{figure}[t]
\centerline{\epsfig{figure=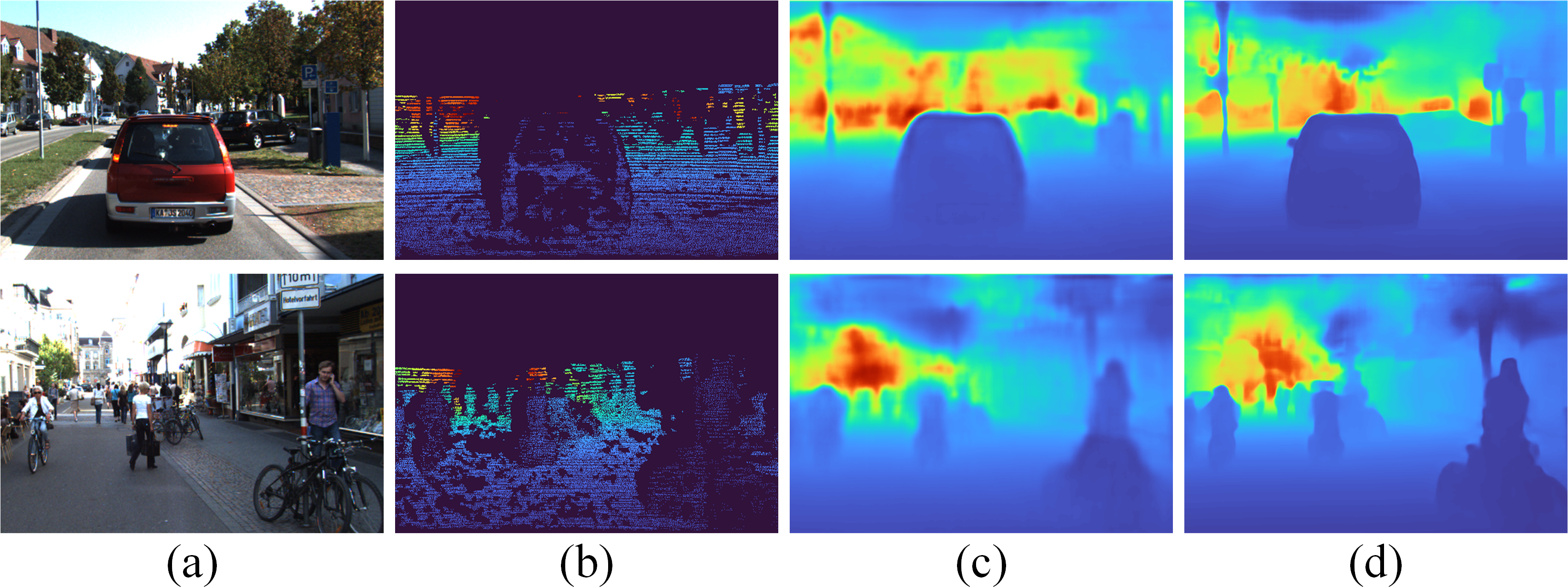,width=8.5cm}}
\vspace{-0.3cm}
\caption{Depth estimation example. (a) Input RGB image; (b) Ground truth depth; (c) Fu et al. \cite{DBLP:conf/cvpr/FuGWBT18/DORN}; (d) Ours.}
\vspace{-0.8cm}
\label{fig 1}
\end{figure}
\begin{figure*}[t]
\includegraphics[width=\textwidth]{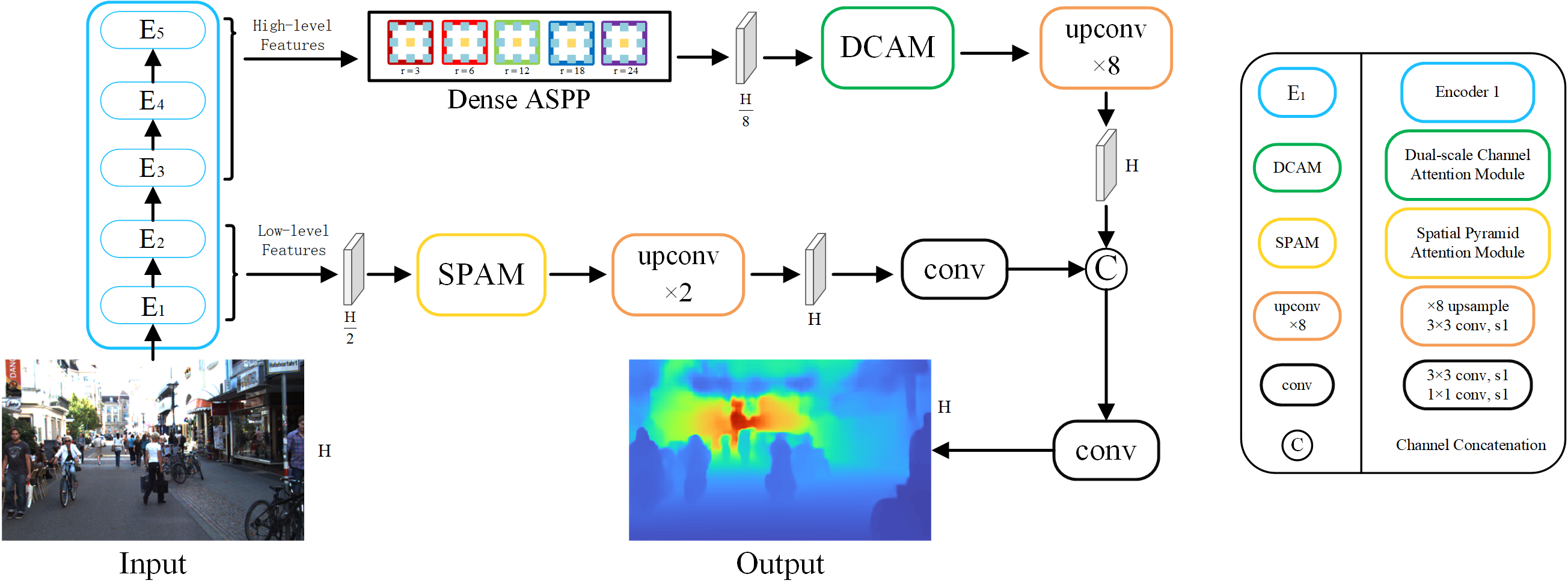}
\vspace{-0.6cm}
\caption{The overview of Pyramid Feature Attention Network. The network is composed of $E_{i}$ (the $i$-th level of encoder), Dense ASPP \cite{DBLP:conf/cvpr/YangYZLY18/DenseASPP}, Dual-scale Channel Attention Module and Spatial Pyramid Attention Module. The high-level features are from $E_{3}$, $E_{4}$ and $E_{5}$. The low-level features are from $E_{1}$ and $E_{2}$.}
\vspace{-0.4cm}
\label{fig 2}
\end{figure*}

In this paper, we propose a novel monocular depth estimation network named Pyramid Feature Attention Network (PFANet). In order to enhance the global structural information in high-level features, we introduced the Dense version of Atrous Spatial Pyramid Pooling (Dense ASPP) \cite{DBLP:conf/cvpr/YangYZLY18/DenseASPP}, which is generally utilized in pixel-level semantic segmentation. Since Dense ASPP applies sparse convolutions with various expansion rates, these convolutions expand receptive field of the high-level features. And then we design Dual-scale Channel Attention Module (DCAM) to aggregate global context and local information at different scales in high-level features. During training process, DCAM assigns larger weight to the channels that play an important role in MDE. Considering the spatial relationship of visual features, we design Spatial Pyramid Attention Module (SPAM) to fuse the attention of multi-scale low-level features. This module improves the detailed local information in the low-level features, which clearer object edge and smoother object surface in prediction depth. Besides, we introduce scale-invariant gradient loss \cite{DBLP:conf/cvpr/UmmenhoferZUMID17/DeMon} to lead the network to learn more detail of object edges. With the above operations, the proposed PFANet can produce good depth maps (see Fig.1 (d)). In summary, our contributions are as follows:

1) We propose a novel Pyramid Feature Attention Network (PFANet) for MDE. For high-level features, we design Dual-scale Channel Attention Module (DCAM) to aggregate global context and local information. For low-level features, we design Spatial Pyramid Attention Module (SPAM) to capture more detailed information.

2) We introduce scale-invariant gradient loss to emphasize the depth discontinuity at different object boundaries and enhance smoothness in homogeneous regions.

3) The proposed method achieves state-of-the-art results on KITTI dataset.

\section{Related Work}
\label{sec:related work}
{\bf Monocular Depth Estimation.} As a pioneering work, Saxena et al. \cite{Saxena2005LearningDF} propose to learn depth from visual cues based on Markov Random Field (MRF). Eigen et al. \cite{DBLP:conf/nips/EigenPF14} introduce deep learning network that make coarse global prediction and refine it with local information, and extend it to a multi-scale network for depth estimation \cite{DBLP:conf/iccv/EigenF15}. Since then, given the success of DCNNs in computer vision, more and more depth estimation networks have been proposed. Laina et al. \cite{DBLP:conf/3dim/LainaRBTN16/FCRN} use a fully convolutional architecture with residual upsampling blocks to tackle the high-dimension regression problem. Jiao et al. \cite{Jiao2018LookDI/lookdeeper} apply semantic segmentation network to assist depth estimation, and propose attention-driven loss that address long-tail distribution of depth values. The latest SOTA network DORN \cite{DBLP:conf/cvpr/FuGWBT18/DORN} models MDE as an ordinal regression problem, to address the increase in error with depth magnitude, via spacing-increasing discretization  strategy.

{\bf Attention Mechanism.} Attention mechanism is derived from human perception, it can selectively focus on the prominent parts to capture useful information in entire scene. Similarly, attention mechanism is also suitable for various computer vision tasks, such as image classification, depth estimation, etc. SENet \cite{DBLP:conf/cvpr/HuSS18/SENet} proposes channel attention module to adaptively recalibrate channel-wise feature responses by explicitly modeling the interdependence between channels. CBAM \cite{Woo2018CBAMCB/CBAM} introduces spatial attention module based on channel attention module, and concatenates two modules for adaptive feature refinement. Wang et al. \cite{DBLP:conf/cvpr/WangWZG20/Hierarchy} design pyramid diverse attention (PDA) to learn multi-scale diverse local representations automatically, leading to network focus on different local patches.
\begin{figure*}[t]
\centering
\includegraphics[width=\textwidth]{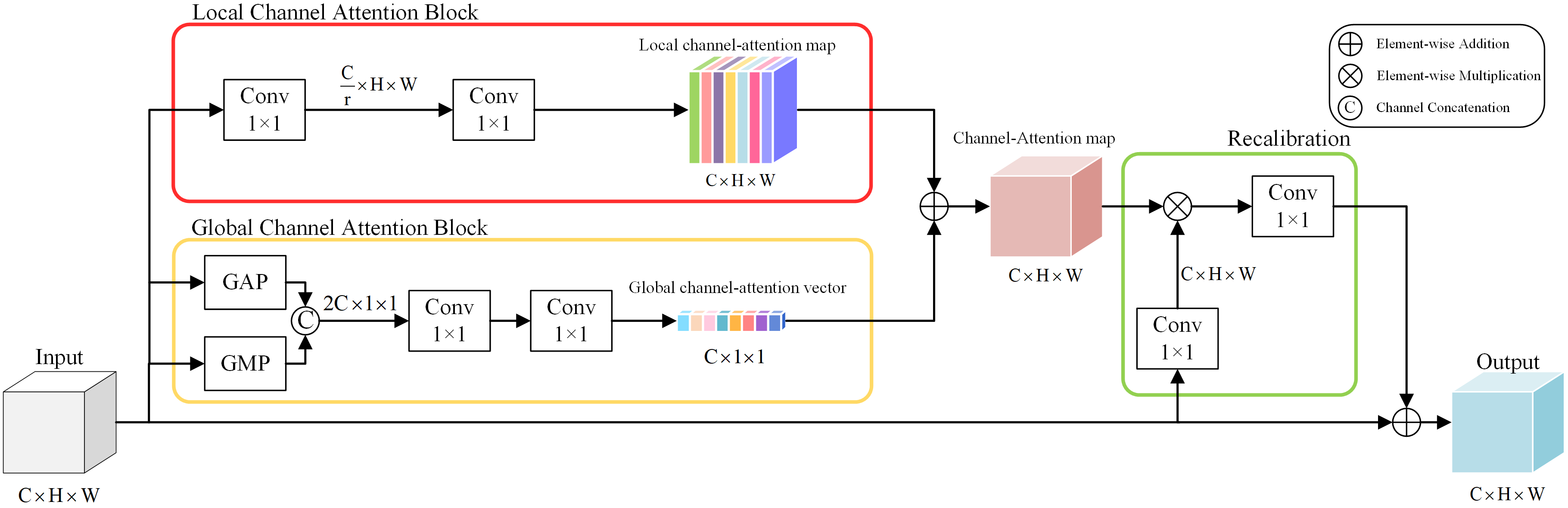}
\vspace{-0.8cm}
\caption{The architecture of Dual-scale Channel Attention Module (DCAM). It consists of two blocks: local channel attention block and global channel attention block. The outputs of two blocks are fused to generate the channel attention map. Recalibration block is utilized to calibrate the channel attention map and further extract useful information for MDE. GAP denotes global average pooling layer. GMP denotes global max pooling layer.}
\vspace{-0.3cm}
\label{fig 3}
\end{figure*}
\section{Our Method}
\subsection{Overview}
In this paper, we propose Pyramid Feature Attention Network based on encoder-decoder architecture. DenseNet-161 \cite{Huang2017DenselyCC/DenseNet} pre-trained on ILSVRC as our encoder. Decoder is composed of Dense ASPP \cite{DBLP:conf/cvpr/YangYZLY18/DenseASPP}, Dual-scale Channel Attention Module, and Spatial Pyramid Attention Module.

The proposed network architecture is shown in Fig.2. Input a single RGB image with resolution \emph{H}. In encoder, the five convolutional blocks \{$E_{1}, E_{2}, E_{3}, E_{4}, E_{5}$\} output feature maps with different resolutions that are \emph{H/2}, \emph{H/4}, \emph{H/8}, \emph{H/16} and \emph{H/32} respectively. The high-level features are from $E_{3}$, $E_{4}$ and $E_{5}$. The low-level features are from $E_{1}$ and $E_{2}$, which upsample to resolution of $E_{2}$. After the backbone network, for high-level features, we apply Dense ASPP module to fuse high-level features and expand the receptive field. This module produces an \emph{H/8} feature map via various dilated convolutional operations. The dilation rates \emph{r} are 3, 6, 12, 18 and 24 respectively. Following Dense ASPP, we place DCAM to extract global context and local information from high-level features. Then, we apply SPAM to capture spatial information at multi-scale from the low-level feature maps. To get the high-level and low-level features with same resolution \emph{H}, we utilize up-convolutional layer, which consists of upsampling operations and a 3×3 convolutional layer. Finally, them are concatenated and fed into the final convolutional layer to get the depth estimation $\widetilde{d}$.

\subsection{Dual-scale Channel Attention Module}
The previous channel attention methods based on the squeeze and excitation \cite{DBLP:conf/cvpr/HuSS18/SENet} capture global context from the feature maps. However, this way ignores local information in features. To aggregate global context and local information simultaneously, we propose Dual-scale Channel Attention Module, as shown in Fig.3. DCAM consists of global channel attention block, local channel attention block and recalibration block. Its core idea is to implement channel attention on different scales.

In global channel attention block, average pooling layer and max pooling layer apply to reduce computational cost. To aggregate global context across channels, the dimension of the input feature maps are fused and reduced to \emph{2C/r} by 1×1 convolution layer, where \emph{C} is the dimension of input $x\in \mathbb{R}^{C\times H\times W}$, \emph{r} is reduction rate. And then this block produces global channel-attention vector $g(x)\in \mathbb{R}^{C\times 1\times 1}$ via another 1×1 convolution layer. Similarly, in local channel attention block, we place two 1×1 convolution layers to extract local information across channels, which generate local channel-attention map $l(x)\in \mathbb{R}^{C\times H\times W}$. Local channel-attention map and global channel-attention vector are fused to channel-attention map $A_{c}(x)\in \mathbb{R}^{C\times H\times W}$, before calibration (see Eq.(1)). Thus, we can effectively employ channel attention information and avoid introducing interference. Finally, we calibrate the original channel-attention map to improve feature representation, and new channel-attention map $\tilde{A}_{c}(x)$ as shown in Eq.(2).
\begin{equation}
    A_{c}(x)=l(x)\otimes g(x)
\end{equation}
\begin{equation}
    \tilde{A}_{c}(x)=h[f(x)\otimes A_{c}(x)]\oplus x
\end{equation}
where \emph{f} and \emph{h} are 1×1 convolutional layers in recalibration block. $\oplus$ denotes element-wise addition. And $\otimes$ denotes element-wise multiplication. Note that each convolutional layer is followed by an activation function ReLU.
\begin{figure}[t]
\centering
\centerline{\epsfig{figure=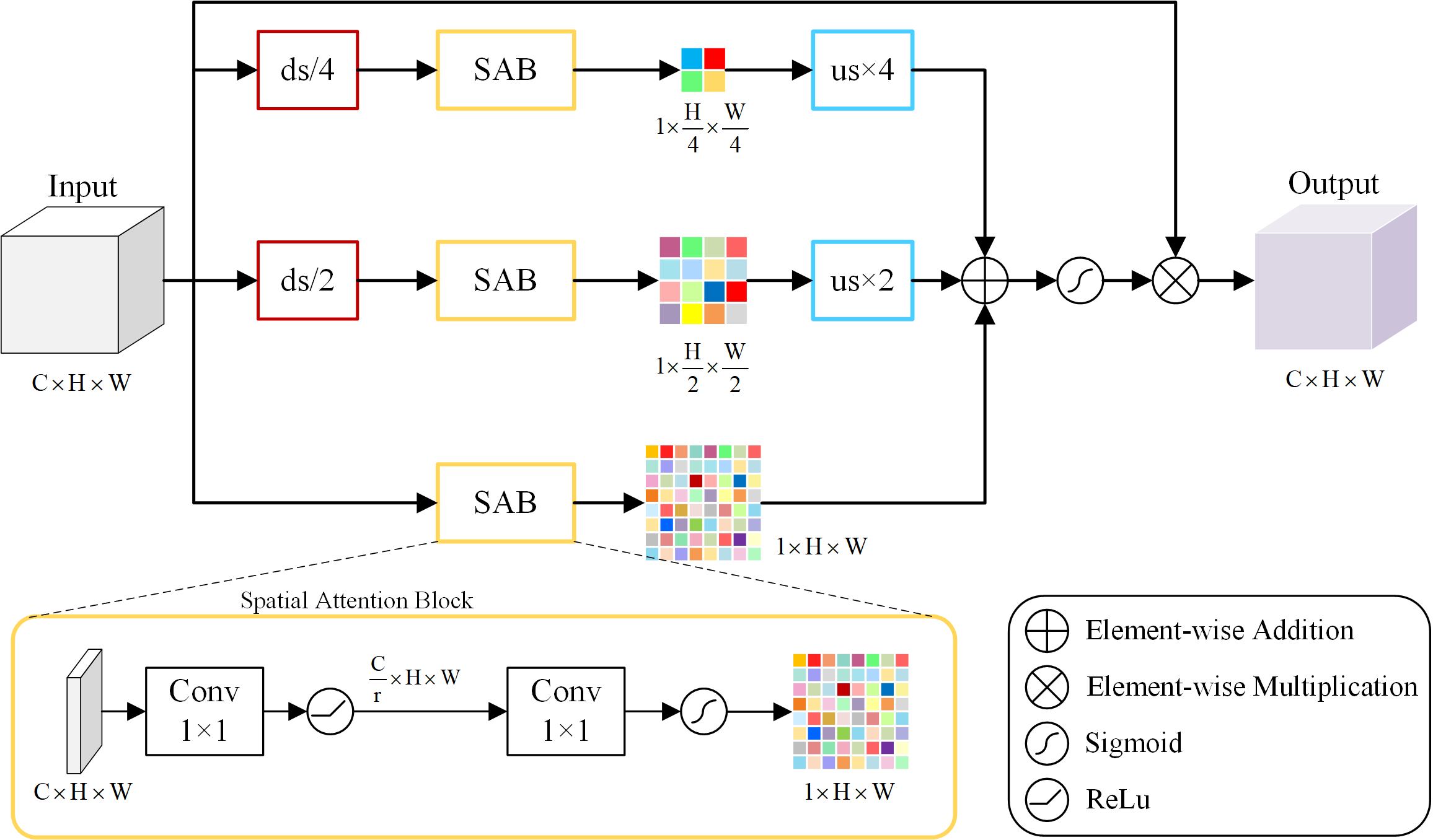,width=8.5cm}}
\vspace{-0.3cm}
\caption{The architecture of Spatial Pyramid Attention Module (SPAM). Ds/4 refers to /4 downsampling operation. Us×4 refers to ×4 upsampling operation. Spatial attention blocks learn the spatial attention map, these three maps form a pyramid structure.}
\vspace{-0.3cm}
\label{fig 4}
\end{figure}

\subsection{Spatial Pyramid Attention Module}
The high-level feature maps always processed by channel attention module, since it can capture channel's dependency. However, this ignores structural information of the feature maps. To extract more local detailed information from the low-level feature maps, we proposed the spatial pyramid attention module, which utilizes the spatial pyramid structure. Fig.4 depicts the paradigm of SPAM. This module contains downsampling operation, spatial attention block and upsampling operation. Suppose the input low-level features $y\in \mathbb{R}^{C\times H\times W}$ via down-sampling operation, get down-sampling feature maps $y_{i}$ \emph{(i=1, 2, 3)}, the resolution is 1, 1/2, 1/4 of the input, respectively. Then spatial attention block learns the spatial attention map $s(y_{i})\in \mathbb{R}^{C\times H\times W}$, as shown in Eq. (3), these three blocks form a pyramid structure. Finally, through upsampling operation, we fuse the multi-scale spatial attention map. The output of SPAM $A_{s}(y)$ can be presented as Eq. (4).
\begin{equation}
    s(y_{i})=\sigma (Conv_{2}(\delta (Conv_{1}(y_{i}))))
\end{equation}
\begin{equation}
    A_{s}(y)=\sigma [s(y_{1})\oplus s(y_{2})\oplus s(y_{3})]\otimes y
\end{equation}
where $Conv_{1}$ and $Conv_{2}$ refer to convolutional layers in spatial attention block. $\delta $ refers to ReLU function, $\sigma $ refers to Sigmoid function.

\subsection{Training Loss}
The loss function to constraint our network contains two terms, i.e., scale-invariant loss (in log space) $L_{d}$ and scale-invariant gradient loss $L_{s}$. We describe in detail each loss item as follows.

{\bf Scale-invariant loss.} Scale invariant loss is proposed in \cite{DBLP:conf/nips/EigenPF14} by Eigen et al., as shown in Eq. (5).
\begin{equation}
    L_{d}(e)=\frac{1}{T}\sum_{i}^{}e_{i}^{2}-\frac{\lambda }{T^{2}}(\sum_{i}^{}e_{i})^{2}
\end{equation}
where $e_{i}=log(\tilde{d}_{i})-log(d_{i})$. $\tilde{d}_{i}$ denotes ground truth of depth. ${d}_{i}$ denotes predicted depth. $\lambda =0.5$. \emph{T} refers to the number of pixels with valid ground truth depth value. By rewritig Eq. (5):
\begin{equation}
   L_{d}(e)=\frac{1}{T}\sum_{i}^{}e_{i}^{2}-\frac{1}{T^{2}}(\sum_{i}^{}e_{i})^{2}+\frac{(1-\lambda )}{T^{2}}(\sum_{i}^{}e_{i})^{2}
\end{equation}
we can think of Eq. (6) as the sum of variances and weighted mean square error in log space. Setting $\lambda =0$ is L2-norm, and setting $\lambda =1$ is scale invariant error. Inspired by \cite{DBLP:journals/corr/abs-1907-10326/BTS}, we set $\lambda =0.85$ in this work to accelerate minimizing the variance of error.

{\bf Scale-invariant gradient loss.} Scale-invariant gradient loss (see Eq. (7)) is defined by \cite{DBLP:conf/cvpr/UmmenhoferZUMID17/DeMon} by Ummenhofer et al., which based on gradient loss. This loss function emphasizes sharpness at object boundaries and increases smoothness with in similar fields. To cover gradients at multi-scale, we utilize 5 different spacings $s\in \left \{1, 2, 4, 8, 16  \right \}$.
\begin{equation}
   L_{g}(g)=\frac{1}{T}\sum_{s\in \left \{ 1, 2, 4, 8, 16 \right \}}\sum_{i,j}\left \| \tilde{g_{s}}(i,j)- g_{s}(i,j) \right \|_{2}
\end{equation}
\begin{equation}
   g_{s}(i,j)=(\frac{d_{i+s,j}-d_{i,j}}{|d_{i+s,j}+d_{i,j}|},\frac{d_{i,j+s}-d_{i,j}}{|d_{i,j+s}+d_{i,j}|})^{T}
\end{equation}
where $\tilde{g_{s}}$ and $g_{s}$ refer to gradient pixel \emph{(i,j)} in predicted depth map and ground truth respectively. $d_{i,j}$ is the depth value of pixel \emph{(i,j)}.

{\bf Total loss.} We find that appropriately scaling the range of loss function can accelerate convergence and improve the final predicted result. Our total loss function is defined as follows:
\begin{equation}
   L_{total}=\alpha \sqrt{L_{d}}+\beta \sqrt{L_{g}}
\end{equation}
where $\alpha $ and $\beta $ are constants we set to 10 and 2 for all experiments. 
\begin{figure}[t]
\centering
\centerline{\epsfig{figure=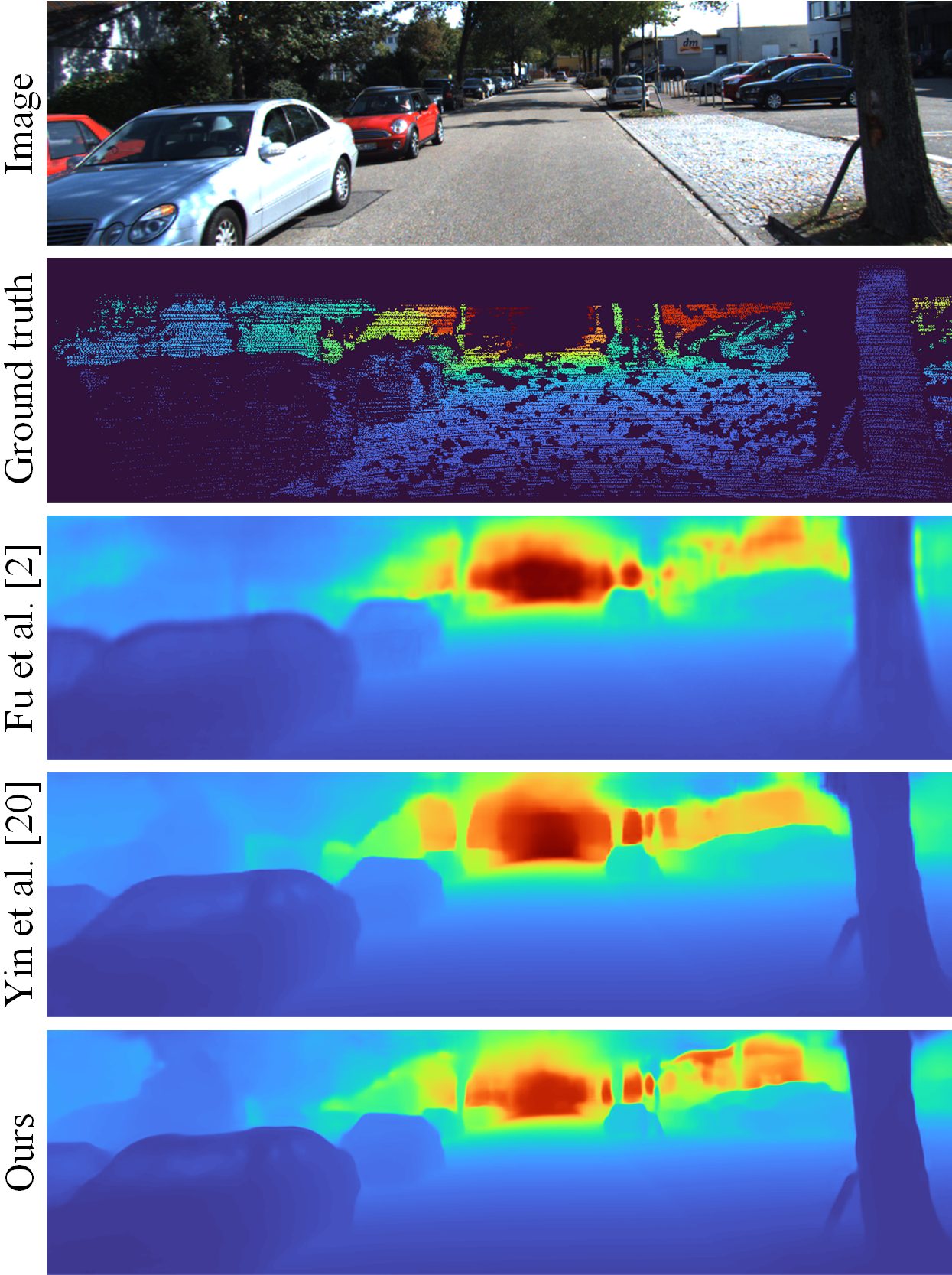,width=8.5cm}}
\vspace{-0.3cm}
\caption{Visualization of the different methods and our proposed method on KITTI dataset.}
\vspace{-0.4cm}
\label{fig 5}
\end{figure}

\begin{table*}[t]
\begin{center}
\caption{Quantitative results on KITTI using Eigen split.} \label{tab1}
\begin{tabular}{|c|c|c|c|c|c|c|c|}
\hline
\multirow{2}{*}{Method} &
  \multicolumn{3}{c|}{Accuracy Metric(higher is better)} &
  \multicolumn{4}{c|}{Error Metric(lower is better)} \\ \cline{2-8} 
 &
  \textbf{$\delta < 1.25$} &
  \textbf{$\delta < 1.25^{2}$} &
  \textbf{$\delta < 1.25^{3}$} &
  \textbf{Abs Rel} &
  \textbf{Sq Rel} &
  \textbf{RMSE} &
  \textbf{RMSE(log)} \\ \hline
Saxena et al. \cite{Saxena2005LearningDF}   & 0.601 & 0.820 & 0.926 & 0.280 & 3.012 & 8.734 & 0.361 \\
Eigen et al. \cite{DBLP:conf/nips/EigenPF14}    & 0.692 & 0.899 & 0.967 & 0.190 & 1.515 & 7.156 & 0.270 \\
Eigen et al. \cite{DBLP:conf/iccv/EigenF15}   & 0.769 & 0.950 & 0.988 & 0.158 & 1.210 & 6.410 & 0.214 \\
Alhashim et al. \cite{DBLP:journals/corr/abs-1812-11941/DenseDepth} & 0.886 & 0.965 & 0.986 & 0.093 & 0.589 & 4.170 & 0.171 \\
Fu et al. \cite{DBLP:conf/cvpr/FuGWBT18/DORN}      & 0.932 & 0.984 & 0.994 & 0.072 & 0.307 & 2.727 & 0.120 \\
Yin et al. \cite{DBLP:conf/iccv/YinLSY19}     & 0.938 & 0.990 & 0.998 & 0.072 & -     & 3.258 & 0.117 \\
Ours &
  \textbf{0.957} &
  \textbf{0.994} &
  \textbf{0.999} &
  \textbf{0.061} &
  \textbf{0.236} &
  \textbf{2.699} &
  \textbf{0.096} \\ \hline
\end{tabular}
\vspace{-0.8cm}
\end{center}
\end{table*}

\begin{table*}[]
\caption{Experimental results using KITTI Eigen split with various backbone networks.} \label{tab2}
\begin{center}
\begin{tabular}{|c|c|c|c|c|c|c|c|c|}
\hline
\multirow{2}{*}{Variant} & \multirow{2}{*}{\#Params} & \multicolumn{3}{c|}{Accuracy Metric(higher is better)} & \multicolumn{4}{c|}{Error Metric(lower is better)} \\ \cline{3-9} 
 &  & \textbf{$\delta < 1.25$} & \textbf{$\delta < 1.25^{2}$} & \multicolumn{1}{c|}{\textbf{$\delta < 1.25^{3}$}} & \multicolumn{1}{c|}{\textbf{Abs Rel}} & \multicolumn{1}{c|}{\textbf{Sq Rel}} & \multicolumn{1}{c|}{\textbf{RMSE}} & \multicolumn{1}{c|}{\textbf{RMSE(log)}} \\ \hline
DenseNet-161 \cite{Huang2017DenselyCC/DenseNet} & 46.6M & 0.955 & 0.993 & \multicolumn{1}{c|}{0.998} & \multicolumn{1}{c|}{0.065} & \multicolumn{1}{c|}{0.251} & \multicolumn{1}{c|}{2.788} & \multicolumn{1}{c|}{\textbf{0.096}} \\
ResNet-101 \cite{DBLP:conf/cvpr/HeZRS16/ResNet} & 68.0M & 0.956 & 0.993 & \multicolumn{1}{c|}{\textbf{0.999}} & \multicolumn{1}{c|}{0.063} & \multicolumn{1}{c|}{0.242} & \multicolumn{1}{c|}{2.721} & \multicolumn{1}{c|}{0.097} \\
ResNext-101 \cite{DBLP:conf/cvpr/XieGDTH17/ResNext} & 112.3M & \textbf{0.957} & \textbf{0.994} & \multicolumn{1}{c|}{\textbf{0.999}} & \multicolumn{1}{c|}{\textbf{0.061}} & \multicolumn{1}{c|}{\textbf{0.236}} & \multicolumn{1}{c|}{\textbf{2.699}} & \multicolumn{1}{c|}{\textbf{0.096}} \\ \hline
\end{tabular}
\vspace{-0.5cm}
\end{center}
\end{table*}

\section{Experiments}
\subsection{KITTI Dataset}
The KITTI dataset \cite{Geiger2013VisionMR/KITTI} consists of 61 outdoor scene images, each with a resolution of 375×1241. Since previous work is based on the training set and test set divided by Eigen et al. \cite{DBLP:conf/nips/EigenPF14}, we also follow it to compare with those works. The training set contains 23488 images from 32 different scenes, and the test set contains 697 images from 29 scenes. The maximum depth of the image in the KITTI dataset is 80.

\subsection{Implementation Details}
We implement the proposed network based on public deep learning framework PyTorch. In training phase, we use Adam optimizer with $\beta _{1}=0.9$, $\beta _{2}=0.999$ and $\varepsilon =10^{-6}$. The learning strategy applies polynomial decay with initial learning rate $lr=10^{-4}$ and power $p=0.9$. We train our network on two NVIDIA TITAN RTX GPU with 24GB memory. Epoch is set to 50 with batch size 32, which applies to all experiments of this work. As the backbone network for encoder, we use ResNet \cite{DBLP:conf/cvpr/HeZRS16/ResNet}, ResNext \cite{DBLP:conf/cvpr/XieGDTH17/ResNext} and DenseNet \cite{Huang2017DenselyCC/DenseNet} pretrained in ILSVRC dataset. Upconvolution in decoder uses the bilinear neighbor upsampling followed by convolutional layer. Downsampling operation and upsampling operation in spatial pyramid attention module utilizes the nearest neighbor method. We set reduction ration 
$r$ is 16. To improve training results, we augment images before input to the network using random rotating, random horizontal flipping, random brightness, contrast and color changing. We randomly crop the image size 352×704 to train the network.

\subsection{Evaluation Metrics}
We quantitatively compare our network with state-of-the-art methods both using the following commonly used metrics:
\newline
$-$ Accuracy with Threshold $t$: $\delta =max(\frac{d_{i}}{\tilde{d_{i}}},\frac{\tilde{d_{i}}}{d_{i}})< t$, for $t\in \left \{ 1.25, 1.25^{2}, 1.25^{3} \right \}$
\newline
$-$ Absolute Relative Error (Abs Rel): $\frac{1}{N}\sum _{i=1}^{N}\frac{\left | \tilde{d_{i}}-d_{i} \right |}{d_{i}}$
\newline
$-$ Squared Relative Error (Sq Rel): $\frac{1}{N}\sum _{i=1}^{N}\frac{\left \| \tilde{d_{i}}-d_{i} \right \|^{2}}{d_{i}}$
\newline
$-$ Root Mean Squared Error (RMSE): $\sqrt{\frac{1}{N} \sum_{i=1}^{N}\left \| \tilde{d_{i}}-d_{i} \right \|^{2}}$
\newline
$-$ Root Mean Squared Error in log space (RMSElog): $\sqrt{\frac{1}{N} \sum_{i=1}^{N}\left \| log(\tilde{d_{i}})- log(d_{i}) \right \|^{2}}$
\newline
where \emph{N} is total number of pixels that the ground truth values are available. $\tilde{d_{i}}$ and $d_{i}$ are predicted depth values and ground truth for pixel \emph{i}.

\subsection{Evaluation Results}
Table 1 shows quantitative results compared with the state-of-the-art methods. Our network far outperforms all existing methods. As shown in Fig. 5, our method shows much more precise object boundaries and much more continuous object surfaces. To prove the effectiveness of our proposed method, we utilize various backbone network as encoder, and keep other settings. Table 2 provides experimental results. And the results show that ResNext-101 achieve state-of-the-art result.

\subsection{Ablation Study}
To investigate the importance of different modules in our method, we conduct the ablation study. It can be seen from Table 3 that the proposed model contains all modules (i.e. DCAM, SPAM, scale-invariant gradient loss) to achieve the best performance, which demonstrates that all modules are necessary to get the best monocular depth estimation result.

\begin{table*}[t]
\caption{Result from the ablation study using KITTI dataset. Baseline: a network composed of only encoder, Dense ASPP, convolution layer and upconvolution layer; L: the network introduce scale-invariant gradient loss. All methods use DenseNet-161 as encoder.} \label{tab3}
\begin{center}
\begin{tabular}{|c|c|c|c|c|c|c|c|}
\hline
\multirow{2}{*}{Method} & \multicolumn{3}{c|}{Accuracy Metric(higher is better)} & \multicolumn{4}{c|}{Error Metric(lower is better)} \\ \cline{2-8} 
 & \textbf{$\delta < 1.25$} & \textbf{$\delta < 1.25^{2}$} & \textbf{$\delta < 1.25^{3}$} & \textbf{Abs Rel} & \textbf{Sq Rel} & \textbf{RMSE} & \textbf{RMSE(log)} \\ \hline
baseline & 0.928 & 0.981 & 0.992 & 0.086 & 0.338 & 3.437 & 0.158 \\
baseline + DCAM & 0.942 & 0.989 & 0.996 & 0.070 & 0.293 & 3.015 & 0.112 \\
baseline + SPAM & 0.945 & 0.990 & 0.997 & 0.068 & 0.253 & 2.841 & 0.106 \\
baseline + DCAM + SPAM & 0.951 & 0.992 & \textbf{0.998} & 0.065 & \textbf{0.251} & 2.810 & 0.098 \\
baseline + DCAM + SPAM + L & \textbf{0.955} & \textbf{0.993} & \textbf{0.998} & \textbf{0.063} & \textbf{0.251} & \textbf{2.788} & \textbf{0.096} \\ \hline
\end{tabular}
\vspace{-0.5cm}
\end{center}
\end{table*}

\section{Conclusion}
In this paper, we present a novel monocular depth estimation network named Pyramid Feature Attention Network to exploit depth information from different levels and address ambiguous object boundaries and discontinuous object surface issues. This network includes two critical modules : Dual-scale Channel Attention Module and Spatial Pyramid Attention Module, which are utilized to improve high-level context features and low-level spatial features, respectively. We also introduce scale-invariant gradient loss for better results. Extensive experimental results on KITTI dataset show that our method outperforms state-of-the-art methods.

\bibliographystyle{IEEEbib}
\bibliography{icme2021template}

\end{document}